\documentclass[11pt]{article}

\usepackage{amsmath, amssymb} \usepackage{latexsym} \usepackage{amsfonts}
\usepackage{color}
\usepackage{verbatim}
\usepackage{graphicx}

\def\n{\noindent}   \def\bop{\noindent {\bf
Proof.}$
\;$ } \def\eop{\hfill
$\Box$ \vspace{0.3 true cm}}

 \def\vt{\vartheta}

\def\P{{\rm P}}

\def\cR{{\mathcal R}}
\def\cN{{\mathcal N}}

 \def\cH{{\mathcal H}}  \def\cL{{\mathcal L}}

\def\n{\noindent}

\def\bop{\noindent {\bf Proof.\ }} 

\def\bR{{\mathbb R}}

   \def\vt{\vartheta}

\def\bop{\noindent {\bf Proof.}$ \;$ } \def\eop{\hfill $\Box$ \vspace{0.5cm}}

\newtheorem{theorem}{Theorem}[section] 
\newtheorem{corollary}[theorem]{Corollary}

\begin{document}

\title{Theoretical Lower Bounds on the Robustness of Deep ReLU Networks}

\author{V\v{e}ra K\r{u}rkov\'{a}\\
Institute of Computer Science of the Czech Academy of Sciences \\
Pod Vod\'{a}renskou v\v{e}\v{z}\'{i}   2, 18207 Prague, Czech Republic}
\date{}

\maketitle

\begin{abstract}
We present a theoretical study of the robustness of parameterized neural networks to random input perturbations. Specifically, we analyze local robustness by quantifying the probability that a random 
$\mathcal{L}_2$-perturbation of a given input results in a correct classification. For deep ReLU networks, we derive lower bounds on local robustness by combining tools from high-dimensional geometry, in particular concentration of measure, with a new characterization of the geometric structure induced by their input-output functions. We prove that each convex polyhedral region in the partition of the input space induced by a ReLU network has at most as many faces as there are network units, regardless of the network depth or architecture. This geometric property serves as the key ingredient in our robustness analysis. Finally, we analyze how local robustness scales with input dimension and characterize the sets of inputs whose neighborhoods are most likely to contain adversarial examples. We show that the width of decision-boundary neighborhoods containing vulnerable points shrinks rapidly as dimension increases and grows only logarithmically with the number of network units. We also discuss the volume of a set of vulnerable points in terms of approximately space-filling shapes of decision boundaries.
\end{abstract}

\section{Introduction}

The reliability of AI systems based on neural networks is crucial for their practical
applicability. In particular, networks should maintain their intended functionality even when
exposed to changes or variations in their inputs. The susceptibility of parameterized networks to input
perturbations has been observed in various cases caused either by corruption of data (due to errors or
noise) or by intentionally crafted perturbations. Szegedy et al. \cite{szal14} initiated 
a study of imperceptible perturbations to test images 
leading to their misclassification. They called such images {\em adversarial} and conjectured that

 {\em ``the set of adversarial negatives
is of extremely low probability, and thus is never (or rarely) observed in the test set,
yet it is dense (much like the rational numbers), and so it is found near virtually every
test case."}

The discovery of neural network vulnerability led to extensive research on algorithms constructing adversarial patterns that networks misclassify despite their indistinguishability from the training patterns
(see \cite{wual24} and references therein), finding minimal adversarial distortion for ReLU networks \cite{weal18},  and methods of defense against adversarial attacks (see, e.g., \cite{yual19, gaal25}) and robustness of ReLU training for simultaneous robustness to all $L_p$ perturbations \cite{cral19}.
Theoretical studies of adversarial patterns have mostly focused on 
characterization of the smallest additive perturbations that can change the outputs  \cite{goal15},  on comparing the influence of the 
worst-case and average perturbations \cite{gaal25}, and influence of Lipschitz continuity on robustness
 \cite{zuku25}.

In this paper, we study the $\cL_2$-robustness of parameterized neural networks (either trained or randomly initialized) from a probabilistic perspective. We introduce the concept of {\em local robustness} at an input $x$ with respect to a precision parameter $r$
defined as the probability that the network classifies correctly an input obtained by applying an additive random perturbation of magnitude $r$ to $x$.  Such perturbed images remain within a Euclidean distance $r$ from the original. For sufficiently small values of $r$, perturbed images are likely to be indistinguishable to the human eye from the training image.

Neural networks typically process data represented as high-dimensional vectors.
 Geometry of high-dimensional spaces exhibits some counterintuitive properties, called the ``blessing
of dimensionality'' \cite{do00,goty18,goal19,vk19}, which make some tasks easier or predictable. These benefits of high-dimension are based on the ``concentration of measure phenomenon" \cite{le51,misc86, gupr24}, which implies, e.g.,  the concentration of the volume and the surface
of high-dimensional spheres \cite{ba97}, the reduction of dimensionality by random projections \cite{joli84}, and the exponential growth of the quasi-orthogonal dimension  \cite{kavk93}. 
High-dimensional geometry also affects functionality and accuracy of neural networks (see e.g., \cite{vksa16,vksa19,sual23,baal23,vksa23}). 

In this work, we investigate how input dimensionality influences the local robustness of parameterized deep ReLU networks. We derive probabilistic lower bounds on their local robustness in $x \in \bR^d$ in terms of the input dimension, the total number of units, and the distance of $x$ from the boundary of the partition induced by a ReLU network on the input space. We show that the lower bounds on robustness at inputs in $\bR^d$ that are not in a close neighborhood of the decision boundary grow rapidly with increasing input dimensionality $d$, while exhibiting only a mild decrease with respect to the total number of units, independently of how these units are distributed across layers. 

Our theoretical analysis is based on characterizing the geometric complexity of the sets of input-output functions of deep ReLU networks, exploiting their piecewise-linear, plane-wave structure.  We prove that a key structural property of the partitions induced by ReLU networks is that each region has only a ``small'' number of faces, bounded above by the total number of network units. By combining this characterization with concentration phenomena in high-dimensional geometry, we derive probabilistic lower bounds on local robustness. Our estimates characterize neighborhoods of decision boundaries in which inputs are more likely to be vulnerable to random $\mathcal{L}_2$ perturbations. We further investigate how the size of these neighborhoods scales with the input dimension and the total number of network units.

The remainder of the paper is organized as follows. Section~\ref{sec:prel} introduces the necessary background, notation, and preliminary concepts. Section~\ref{sec:prob} develops a probabilistic framework for analyzing robustness to misclassification caused by random additive perturbations. Within this framework, we first illustrate the fundamental role of data dimensionality using the example of classification by a single Heaviside perceptron, and then extend the probabilistic misclassification bounds to shallow perceptron networks. Section~\ref{sec:deep} includes structural analysis of sets of input-output functions of deep ReLU networks and exploits it to derive our main results on their robustness and analyze them. Finally, Section~\ref{sec:disc} concludes with a summary and discussion of the consequences of our results for the hypothesis of Szegedy et. al \cite{szal14} that motivated our work.

\section{Preliminaries} \label{sec:prel}

{\em Feedforward networks} compute parameterized families of input-output (I/O) functions
determined by directed acyclic graphs, where nodes represent computational units and edges represent connections between them. 
In  {\em multilayer networks}, units are arranged in layers, inputs in $X \subset \bR^d$ are viewed as the layer $0$,
layers $l=1, \dots, L-1$ are called {\em hidden layers}, and the number of layers $L$ is called
the {\em network depth}. In this paper, we focus on networks where {\em connections to any layer
are only from the preceding layer}, and where the {\em $L$-th layer contains only one output unit}.

Typical computational units  apply a fixed {\em activation function} $\psi: \bR
\to \bR$ to {\em affine
functions with varying parameters}. The units compute functions of the form
$$\psi(v \cdot . +b): \bR^d \to \bR,$$

\n where $v \in \bR^d$ is called a {\em weight vector}, $b \in \bR$ a {\em bias}, and
 $$v \cdot x = \sum_{i=1}^d v_i x_i$$
\n denotes the {\em scalar product}.
These units have been called {\em perceptrons} because their form was inspired by a naive
model of a neuron with hard
(Heaviside, signum) or soft (hyperbolic tangent, logistic sigmoid) threshold functions
as activations. Currently, the most popular activation function is the rectified linear function
(ReLU) function due to its suitability for gradient-based learning.
We denote by $\vt: \bR \to \{0,1\}$
the { \em Heaviside function}
$$\vt(t) =0 \quad {\rm for} \quad t \leq 0 \quad \vt(t) =1 \quad {\rm for} \quad t > 0,$$
\n by
$\rho: \bR \to \bR$
the {\em rectified linear function}
$$\rho(t) =0 \quad {\rm for} \quad t \leq 0 \quad \rho(t) =t \quad {\rm for} \quad t > 0,$$

\n and for  $v \in S^{d-1}$ and $b \in \bR$, by

$$H_{v,b} :=\{ x \in \bR^d \, | \, v \cdot x +b = 0\}$$

\n the {\em hyperplane} determined by $v$ and $b$.

For our study, it is essential that functions computed by perceptrons with any activation function have the form of {\em plane waves}. They are constant on hyperplanes
parallel to $H_{v,b}$. We denote
$$H_{v,b}^-:=\{ x \in \bR^d \, | \, v \cdot x +b \leq 0\}$$
$$H_{v,b}^+:=\{ x \in \bR^d \, | \, v
\cdot x +b > 0\}$$
\n the two {\em half-spaces} separated by $H_{v,b}$. Note that we define the half-spaces
in such a way that the hyperplane $H_{v,b}$ is contained in the half-space $H^-_{v,b}$ , so $H^-_{v,b}$ is closed and $H^+_{v,b}$ is open. The reason is that the function $\rho$ is equal to zero on the half-line $(\infty,0]$.
Thus $\bR^d = H^+_{v,b} \cup H^-_{v,b}$ is a genuine partition.

For $x \in \bR^d$,  we denote by $p_{H_{v,b}}(x)$ the
{\em orthogonal
projection} of $x$ onto $H_{v,b}$ and
for $H \subset \bR^d$ and $x \in \bR^d$, 
$$\| x - H \|_2: = \inf_{z \in H} \|
x-z\|_2$$ \n  the {\em Euclidean distance}
of $x$ from $H$, by $S_r^{d-1}(x)$ the {\em sphere} in $\bR^d$ of radius $r$ centered at $x$,
and by $S^{d-1} :=
S_1^{d-1}(0)$ the {\em unit
sphere} centered at $0$.

We denote by
\begin{equation}
\cN:= \cN( v_{l,j}, b_{l,j}, j=1, \dots, n_l, l=1, \dots, L-1, v_L, b_L)
\label{eq:N}
\end{equation}

\n the {\em  parameterized $L$-layer network}, where each of the layers $l=1, \dots, L-1$
contains $n_l$ perceptrons
with ReLU activations and parameters $v_{l,j},b_{l,j}$, and the layer $L$ contains one
Heaviside perceptron computing $\vt(v_L \cdot y_{L-1} + b_L)$, 
where $y_{L-1}= (y_{L-1,1}, \dots, y_{L-1,n_{L-1}} ) \in \bR^{n_{L-1}}$
is the vector of outputs of
$n_{L-1}$ units in the $(L-1)$-st layer,  and by  $$f_{\cN}: \bR^d \to \{0,1\}$$ 
\n  the {\em  input-output (I/O) function} of the network $\cN$.

\section{Probabilistic approach to robustness}
 \label{sec:prob}

For a parameterized network $\cN$ (with parameters obtained by training or set randomly) that computes
 an I/O function $f_{\cN}$, and for an input $x \in \bR^d$, we study the local robustness of $\cN$ at $x$ in probabilistic terms.
We assume that $x$ is modified by an additive random perturbation $\xi$ of a given size $r$, formally, by $\xi$ drawn from the sphere $S_r^{d-1}(x)$ according to the uniform probability $\P$.  The Euclidean distance between $x$ and $x+\xi$ is $r$, and therefore for small $r$,  in the case of images, these two inputs might be indistinguishable.

We measure robustness of a parameterized network $\cN$ by the probability that
$f_{\cN}$ maps  $x+\xi$ to the same class as $x$. Formally, we define the
{\em  local robustness of  $\cN$ at $x$ to perturbations of the size $r$} as

\begin{equation}
\cR(f_{\cN},x,r):= \P[ \, f_{\cN}(x+\xi)= f_{\cN}(x)\, ].
\label{eq:R}
\end{equation}

To gain insight into the robustness of networks computing plane waves,
we begin with the simplest case: a single Heaviside perceptron.
Let $$f(x) := \vt(v \cdot x +b)$$ \n be the I/O
function of a parameterized
perceptron with parameters $v \in \bR^d$ and $b \in \bR$, and $X \subset \bR^d$ be the
domain of its inputs. 

Our probabilistic lower bound on its robustness relies on a counterintuitive geometric phenomenon in high dimensions: the fraction of the $d$-dimensional sphere occupied by a polar cap of fixed angular radius decreases exponentially fast with 
$d$ (see Fig.~1).

\begin{figure}[ht]
\centerline{\includegraphics[width=5cm]{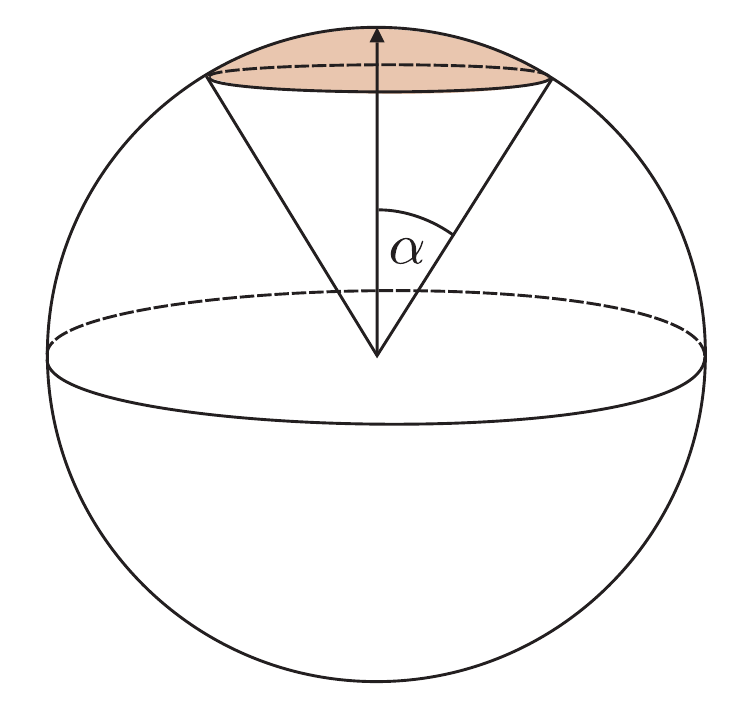}}
\label{fig:capsof}\caption{Spherical polar cap.}
\end{figure}

\begin{theorem}
 Let $f: \bR^d \to \{0,1\}$, defined as $f(x)=\vt(v \cdot x +b)$, be an I/O function
of a perceptron with the Heaviside activation $\vt$, $r>0$ be a size of a perturbation. Then for every   
 $x \in\bR^d$, $a(x) = \| p(x) - x\|_2$, where $p(x)$ is the orthogonal projection of $x$
on the hyperplane $H_{v,b}$,and $\xi$ drawn uniformly randomly from  $S_r^{d-1}(x)$, we have:\\
(i) for $0 < a(x) \leq r$,
$$\cR(f,x,r):=\P [f(x) = f(x + \xi)] >1 - e^{-\frac{a(x)^2 d}{2r^2}};$$
\n (ii) for $a(x)=0$, $$\cR(f,x,r) :=\P[(f(x) =f(x + \xi)] =1/2;$$
\n (iii) for $r < a(x)$,  $$\cR(f,x,r) := \P [f(x) = f(x + \xi)]=1.$$
\label{th:oneunit}
\end{theorem}

\bop
For $y \in S^{d-1}$ and $\alpha \in (0,\pi/2)$, let $$C(y,\alpha) := \{ z \in S^{d-1} \, |
\, z \cdot y  \geq \cos \alpha \}$$
\n denotes the {\em spherical polar cap} formed by the vectors from the sphere
$S^{d-1}$ with the angular distance to $y$ at most $\alpha$ (see Fig.~1). 
Let $\mu$ denote the
normalized Lebesgue measure on $S^{d-1}$, then for every $y \in S^{d-1}$ and every
$\alpha \in (0, \pi/2)$, the polar cap $C(y,\alpha)$ satisfies

\begin{equation}
\mu(C(y, \alpha)) \leq e^{-\frac{(\cos \alpha)^2 \, d}{2}}
 \label{eq:polar}
 \end{equation}

\n  (see, e.g., \cite[Lemma 2.2., p.11]{ba97}).

Rescaling and translating
$S_r^{d-1}(x)$ to $S^{d-1}$, we obtain from (\ref{eq:polar}) for every $y \in S^{d-1}_r(x)$,

\begin{equation}
\mu_r \{ \xi \in S_r^{d-1}(x) \, \Big |  \, | (x+\xi) \cdot y | \geq \cos \alpha \} \leq
e^{-\frac{(\cos \alpha)^2 \, d}{2}},
\label{eq:xpolar}
 \end{equation}

\n where $\mu_r$ denotes the normalized Lebesgue measure on
$S_r^{d-1}(x)$ .

\begin{figure}[ht]
 \centerline{\includegraphics[width=6cm]{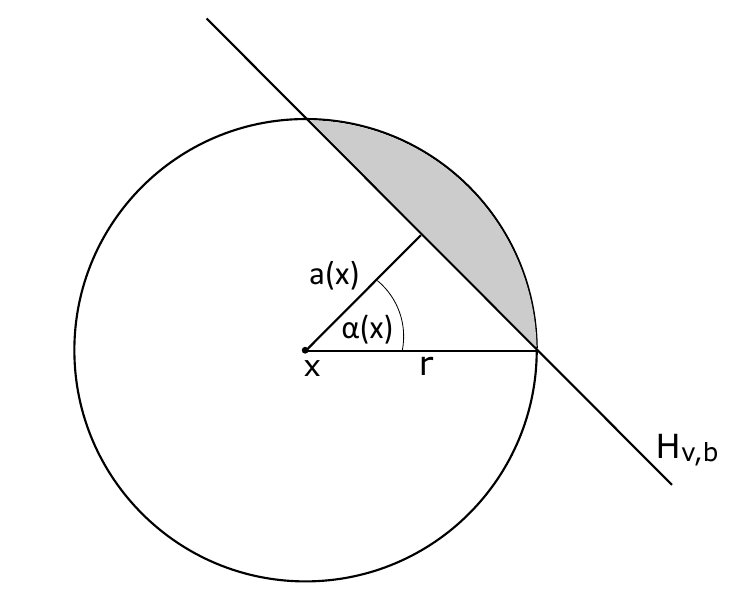}} 
\label{fig:oneunitx}
\caption{Area of perturbed patterns separated from $x$ by a hyperplane.} \end{figure}

(i) If $0< a(x) <r$, then the perturbed input $x+\xi$ is separated from $x$ by the
hyperplane  $H_{v,b}$ if it is in the
polar cap on $S_r^{d-1}(x)$ centered at $y = x+ r \frac{p(x)-x}{a(x)}$ with the angle $\alpha(x)$ such that
$\cos \alpha(x) = \frac{a(x)}{r}$  
(see Fig. 2).
 Thus from (\ref{eq:xpolar}), we get for $\xi$ drawn uniformly from $S^{d-1}_r(x)$

$$\P[ \xi \in S_r^{d-1}(x)\,
| \, f(x+\xi) \neq f(x) ] \leq e^{-\frac{a(x)^2 \, d}{2r^2}}.$$

\n (ii) If $a(x)=0$, then $x \in H_{v,b}$ and thus the probability $\P[x+\xi \in H_{v,b}^-]
=\P[x+\xi \in H_{v,b}^+] =1/2$.

\n  (iii) If $r < a(x) $, then both $x$ and $x+\xi$ are in the same open half-space separated by $H_{v,b}$.
\eop

Theorem \ref{th:oneunit} analyses the local robustness at inputs in dependence on their distance from the hyperplane $H_{v,b}$, the input dimension $d$, and perturbation size $r$.
The  cases (ii), when the original input $x$ is farther from $H_{v,b}$ than the size of perturbation $r$, and  (iii),  when its distance from $H_{v,b}$
is equal to $r$,  are obvious. In these two cases, the local robustness at $x$ does not depend on the input dimension $d$.

Only the case (i),  when the distance from the hyperplane $H_{v,b}$
 is smaller than  $r$, is substantive. 
The probability that a Heaviside perceptron correctly classifies a slightly 
perturbed pattern - lying at Euclidean distance $r$ from the original pattern -
is at least 

\begin{equation}
1- e^{-\frac{a(x)^2 \, d}{2 r^2}}.
\label{eq:lb}
\end{equation}

\n Thus the larger the  quantity $a(x)^2 d$ relative to $2 r^2$, the more robust the network is at $x$.
The lower bound~(\ref{eq:lb}) demonstrates  the beneficial effect of input dimensionality on the robustness of a Heaviside perceptron. 

As a straightforward corollary of Theorem \ref{th:oneunit}, we obtain the following
probabilistic bound
on the robustness of a shallow network with $n$ Heaviside perceptrons.

\begin{corollary}
Let $f(x)=\sum_{i=1}^n w_i\vt(v_i \cdot x +b_i): \bR^d \to \bR$ , where $v_i \in \bR^d$,
$b_i \in \bR$, $i=1, \dots, n$,
be an I/O function of a parameterized shallow network with $n$ Heaviside perceptrons,
$H=\bigcup_{i=1}^n H_{v_i, b_i}$,   and $r  >0$ be a size of a perturbation. 
Then for every $x \in \bR^d \setminus H$ and $\xi$ drawn uniformly randomly from $S_r^{d-1}(x)$, we have
$$\cR(f,x,r):=  \P [ f(x ) = f(x+\xi)] \geq 1- n \, e^{-\frac{ a(x)^2 \, d}{2r^2}},$$
\n where $a(x)= \min \{a_i(x) \, | \, i=1, \dots, n\}$, $a_i(x) = \| x -p_i(x)\|_2$, and 
$p_i(x)$ is the orthogonal projection of
$x$ on $H_{v_i,b_i}$ for $i=1, \dots, n$
\label{co:shall}
\end{corollary}

\begin{figure}[ht]
 \centerline{\includegraphics[width=8cm]{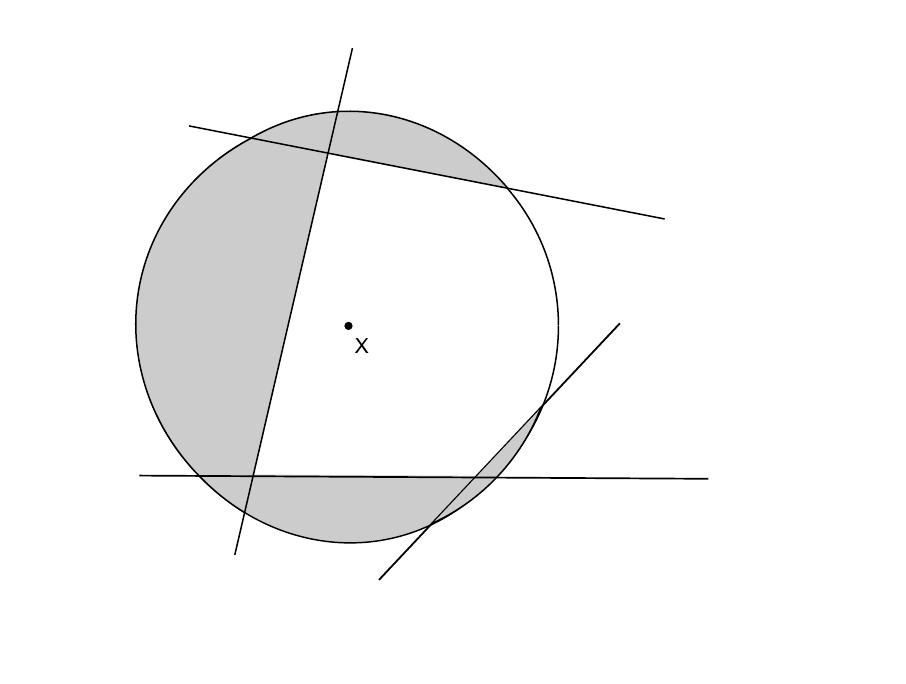}}
 \caption{Robustness of a
shallow Heaviside perceptron network}
\label{fig:shall}
 \end{figure}

Corollary \ref{co:shall} shows that the local robustness at almost all inputs in $\bR^d$ (up to a set $H$ of the Lebesgue measure zero) of a shallow Heaviside perceptron network increases rapidly with the input dimension, provided that the number of network units $n$ does not dominate the term
$e^{-\frac{a(x)^2 \, d}{2r^2}}$ (see Fig. \ref{fig:shall}). 
For example, this condition is satisfied when the number of network units grows polynomially with $d$.

\section{Robustness of deep ReLU networks}
\label{sec:deep}

To extend our probabilistic estimates of local robustness to deep ReLU networks, 
we first show that their I/O functions partition the input domain into polyhedral regions (or convex in case of unbounded ones), each with only a ``small'' number of faces. 
The following theorem establishes that the number of faces bounding any individual region is at most the total number of network units. In contrast to the number of regions, which may grow exponentially with network depth, this upper bound holds regardless of network depth and unit arrangement across layers.

\begin{theorem}
Let $\, \cN( v_{l,j}, b_{l,j}, j=1, \dots, n_l, l=1, \dots, L-1, v_L, b_L)$ be a parameterized
network with $L$ layers, where the layers $l=1, \dots, L-1$
contain $n_l$ ReLU units with parameters $v_{l,j},b_{l,j}$, $j=1, \dots, n_l$, $n_L=1$,
the single unit in the $L$-th layer is the Heaviside
perceptron with parameters $v_L, b_L$, and  $n=\sum_{l=1}^{L} n_l$ denotes the total number of the network units. Then
the I/O function $f_{\cN}: \bR^d \to \{0,1\}$ of $\cN$ is piecewise constant on a
partition of $\, \bR^d$ into convex subsets, such that  each of them
has at most $n$ faces, and each face is contained in a hyperplane.
\label{th:ReLU}
\end{theorem}

\bop
For each layer $l=1, \dots, L-1$, let $f_l: \bR^d \to \bR^{n_l}$ denote the function
$f_l: \bR^d \to \bR^{n_l}$ defined as
$$f_l(x)= y_l(x)=(y_{l,1}(x), \dots, y_{l,n_l}(x)) \in \bR^{n_l},$$
\n  where $y_l(x)$ is the output vector of the $l$-th layer.

To obtain the partition, for each $x^* \in \bR^d$, we construct a convex set $D_L(x^*)$, on which the I/O function
 $f_{\cN}$ is constant, and then
identify those sets $D_L(x^*)$ and $D_L({\bar x}^*)$ that coincide.

We built the set  $D_L(x^*)$ recursively from a sequence of nested convex
subsets
$\{D_l(x^*) \, | \, l=1, \dots, L\}$ of
$\bR^d$, that all contain $x^*$. 
We construct them together with $L$ families of hyperplanes
$$\cH_l =\{H_{w_{k,j}, c_{k,j}} \, | \, k=1, \dots, l, j=1,\dots, n_k \}, \quad l=1, \dots L,$$ 
\n such that the faces of each of $D_l(x^*)$ , $l=1,
\dots, L$, are formed by
subsets of the hyperplanes $\{H_{w_{k,j}, c_{k,j}} \, | \, k=1, \dots, l, j=1,\dots,
n_l\}$.

We start with the first layer $l=1$. We set $w_{1,j} = v_{1,j}$ and $c_{1,j} = b_{1,j}$ for all $j=1, \dots, n_1$
and define
$$D_1(x^*)= \bigcap_{j=1}^{n_1} H_{w_{1,j},c_{1,j}}^{\iota_1},$$
\n where $\iota_1$ is either $+$ or $-$
depending on whether $x^* \in H_{ w_{1,j},c_{1,j} }^+$ or
$x^* \in H_{w_{1,j},c_{1,j} }^-$.
The function $f_1: \bR^d \to \bR^{n_1}$, that maps input $x \in \bR^d$ to the output
vector of the $1$-st
hidden layer, satisfies for every $j=1, \dots, n_1$,
$$f_1(x)_j = y_{1,j}(x)=\rho(w_{1,j} \cdot x + c_{1,j}).$$
\n Denoting $$J_1^+ = \{ j \in \{1, \dots, n_1\} \, | \, x^* \in H_{w_{1,j},c_{1,j}}^+ \}$$
\n and
$$J_1^-  = \{ j \in \{1, \dots, n_1\} \, | \, x^* \in H_{w_{1,j},c_{1,j}}^- \},$$ 

\n we get for every $x \in D_1(x^*,)$

$$f_1(x)_j= y_{1,j}(x)  = w_{1,j} \cdot x + c_{1,j} \quad {\rm  when} \quad   j\in J_1^+$$

$${\rm and} \quad f_1(x)_j = y_{1,j}(x)=0  \quad {\rm when} \quad  j \in J_1^-.$$

\n So,  $D_1(x^*)$ is convex, its boundaries are formed by
subsets of at most $n_1$ hyperplanes $H_{w_{1,j},c_{1,j}}$,
$j=1, \dots, n_1$, and on $D_1(x^*)$, the functions $f_1(x)_j= y_{1,j}(x)$, $j=1, \dots,
n_1$, are affine or equal to zero.

Assuming  that we have $D_{l-1}(x^*)$ together with the family of hyperplanes
 $\cH_{l-1}= \{ H_{w_{l-1,j},c_{l-1,j}}$, $j=1, \dots, n_{l-1} \}$,
we construct  a family of hyperplanes  $\cH_l = \{H_{w_{l,k},c_{l,k}}$, $k=1, \dots, n_l\}$
and $D_l(x^*)$. The function $f_l : \bR^d \to \bR^{n_l}$, that maps $x \in \bR^d$ to the
output vector

$$f_l(x) =y_l(x) =(y_{l,1}(x), \dots, y_{l,n_l}(x)) \in \bR^{n_l}$$

\n  of the $l$-th hidden layer, satisfies for every $k=1, \dots, n_l$

$$f_l(x)_k= y_{l,k}(x) = \rho \Bigl  ( \, v_{l,k} \cdot y_{l-1}(x) +b_{l,k} \, \Bigr ) =
\rho \Bigl ( \sum_{j=1}^{n_{l-1}} v_{l,k,j} y_{l-1,j} + b_{l,k} \Bigr )=$$

$$ \rho \Bigl ( \sum_{j\in J^+_{l-1}} \left ( v_{2,k,j} w_{l-1,j} \cdot x + c_{l-1,j}
\right ) +
 \sum_{j \in J^-_{l-1}} c_{l-1,j}  + b_{2,k} \Bigr ).$$

\n Setting $w_{l,k} = \sum_{j\in J^+_{l-1}} v_{l,k,j}w_{l-1,j}$ and $c_{l,k} =\sum_{j\in J^-_{l-1}} c_{l-1,j}
+ b_{l,k}$,
we get

$$y_{l,k}(x) =   \rho( w_{l,k}  \cdot x + c_{l,k}).$$

\n We define

 $$D_l(x^*) = D_{l-1}(x^*)  \cap \bigcap_{j=1}^{n_l} H_{w_{l,k},c_{l,k}}^{\iota_1},$$

\n where $\iota_1$ is either $+$ or $-$
depending on whether $x^* \in H_{ w_{l,k},c_{l,k} }^+$ or
$x^* \in H_{w_{l,k},c_{l,k} }^-$.
Denoting $$J_l^+ = \{ k \in \{1, \dots, n_l\} \, | \, x^* \in H_{w_{l,k},c_{l,k}}^+ \}$$ 
\n and
$$J_l^- = \{ k \in \{1, \dots,
n_l\} \, | \, x^* \in H_{w_{l,k},c_{l,k}}^- \},$$ 
\n we get
for all $x \in D_l(x^*)$,

$$f_l(x)_k= y_{l,k}(x)  = w_{l,k} \cdot x + c_{l,k} \quad  {\rm when}  \quad k\in J_l^+$$

$$f_l(x)_k = y_{l,k}(x)=0 \quad  {\rm when}  \quad k \in J_l^-.$$

\n So,  $D_l(x^*)$ is convex, its faces are formed by
subsets of at most $n_1 +\dots + n_l$ hyperplanes $H_{w_{p,k},b_{p,k}}$,
 $p=1, \dots, l$, $k=1, \dots, n_l$, and the components \\
 $f_l(. )_k= y_{l,k}(x) : \bR^d \to \bR$ of the
function $f_l: \bR^d \to \bR^{n_l}$
that maps $x \in \bR^d$ to the ouput vector $y_l= (y_{l,1}, \dots, y_{l,n_l})$ are  affine or constant zero.

The  I/O function $f_{\cN}: \bR^d \to \{0,1\}$
satisfies
$$f_{\cN}(x)= f_L(x) =\vt(v_L \cdot y_{L-1}(x) + c_L),$$
\n where $y_{L-1}(x) =(y_{L-1,1}(x), \dots, y_{L-1,n_{L-1}}(x)) \in \bR^{n_{L-1}}$.
$D_{L-1}(x^*)$ was constructed in such a way, that for all $x \in D_{L-1}(x^*)$,  we have
$$y_{L-1,s} = w_{L-1,s} \cdot  x + c_{L-1,s} \quad {\rm when} \;s \in J_{L-1}^+$$

$${\rm and } \quad  y_{L-1,s}(x) =0  \quad {\rm when} \; s \in J_{L-1}^-.$$

\n Thus for  $x \in D_{L-1}(x^*)$
$$f_{\cN}(x) = \vt \left  ( \sum_{s=1}^{n_{L-1}} v_{L, s} y_{L-1,s}(x) + b_L)  \right ) =$$

$$ \vt   \left ( \sum_{s \in J^+_{L-1}} (v_{L,s} w_{L-1,s} \cdot x) + \sum_{s \in J^-_{L-1}} c_{L-1,s} +
b_L \right ) .$$

\n Setting $w_L = \sum_{s \in J^+_{L-1}} v_{L,s} w_{L-1,s}$
and $c_L = \sum_{s \in J^-_{L-1}}  c_{L-1,s} + b_L$,
we define

 $$D_L(x^*) = D_{L-1}(x^*)  \cap H_{w_L,c_L}^{\iota},$$

\n where $\iota$ is either $+$ or $-$
depending on whether $x^* \in H_{w_L,c_L }^+$ or
$x^* \in H_{w_L,c_L}^-$.
As $D_L(x^*)$ is either a subset of the positive or the negative half-space induced by the
hyperplane $H_{w_L,c_L}$,
the I/O function $f_{\cN}$ is constant on $D_L(x^*)$
(it is either $0$ or $1$ depending on $\iota$).

It follows from the construction that for each pair $x^*$ and ${\bar x}^*$, $D_L(x^*)$ and $D_L({\bar x}^*)$ are either disjoint or coincide. Identifying coincident regions, we get the partition of $\bR^d$.
\eop

Theorem \ref{th:ReLU} characterizes an I/O function of a parameterized deep ReLU network with a single Heaviside output as a piecewise constant function defined on a partition of $\mathbb{R}^d$
 into convex regions, where each region has at most as many faces as the total number of units in the network. 
This result on bounded complexity property of each individual region measured by its number of faces  is the key feature exploited in the next theorem.

\begin{theorem}
Let $\,\cN( v_{l,j}, b_{l,j}, j=1, \dots, n_l, l=1, \dots, L-1, v_L, b_L)$ be a parameterized
network with $L$ layers, where the layers $l=1, \dots, L-1$
contain $n_l$ ReLU units with parameters $v_{l,j},b_{l,j}$, $j=1, \dots, n_l$, $n_L=1$ and
the single unit in the $L$-th layer is the Heaviside
perceptron with parameters $v_L, b_L$. Then there exists a subset $H$ of $\, \bR^d$ in the
form of a union of subsets of hyperplanes, such that the I/O function $f_{\cN}:\bR^d \to \{0,1\}$ of $\cN$
satisfies for
every $x\in \bR^d \setminus H$, every size of perturbation $r>0$, and every $\xi$ drawn from $S_r^{d-1}(x)$ according to the uniform probability $\P$,
$$\cR(f_{\cN}, x,r) : = \P \Bigl [ f_{\cN}(x) = f_{\cN}(x+\xi) \Bigr ] \geq 1 - n \, e^{-\frac{a(x)^2 \, d}{2r^2}},$$
\n where $n:= \sum_{l=1}^{L} n_l$ is the total number of network units and $a(x) := \| x-
H\|_2$.
\label{th:deep}
\end{theorem}

\begin{figure}[ht]
 \centerline{\includegraphics[width=6cm]{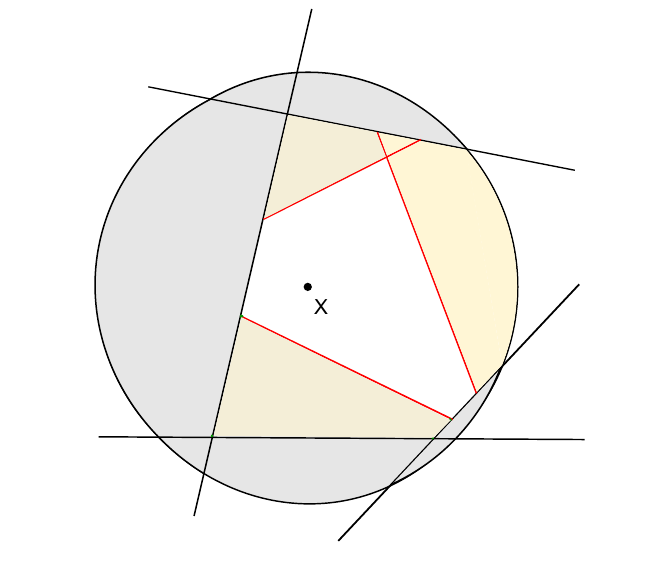}}
 \label{fig:domain2}
\caption{Probability of a misclassification by a deep ReLU
network} \end{figure}

\bop
By Theorem \ref{th:ReLU}, there exists a partition of $\bR^d$ into convex sets, each of
which has at most $n$ faces formed
by subsets of hyperplanes. Let $H$ be the union of faces of these convex sets.
For each $x^* \in \bR^d \setminus H$, $f_{\cN}$ is constant on the set $D_L(x^*)$ from the
partition constructed in the proof of Theorem
\ref{th:ReLU} (see Fig. \ref{fig:domain2}).  On each of these sets, $f_{\cN}$  is equal either to $0$ or $1$.
For a fixed $x^*$, let $H^*_1, \dots, H^*_m$ be the hyperplanes, the
parts of which form faces of $D_L(x^*)$.  By Theorem \ref{th:ReLU}, $m\leq n$.

Using properties of high-dimensional spheres as in the proof of Theorem \ref{th:oneunit}, we get for each $x$ in the interior of $D_L(x^*) $ and for each $i=1, \dots, m$ and  $a_i:= \| x - H^*_i\|_2$ the upper bound 
$$e^{-\frac{a_i(x)^2 \, d}{2r^2}},$$ 
\n on 
 the  probability that $x+\xi$ is separated from $x$ by $H^*_i$.
As  for all $i=1, \dots, m$, $a(x) =  \| x - H\|_2 \leq \|x - H^*_i\|_2$,
we get $$\P[ x+\xi \notin D_L(x^*)] \leq n \, e^{-\frac{a(x)^2 \, d} {2r^2}}.$$ \eop

Note that the bound from Theorem~\ref{th:deep} is conservative in two ways: adjacent regions can have the same label, and $a(x)$ only roughly estimates  $a_i$. 
This bound is non-vacuous for  inputs $x \in \mathbb{R}^d$ satisfying
$$ \frac{a(x)^2 d}{2r^2} >  \ln n  \Longleftrightarrow a(x) > r \sqrt{\frac{2 \ln n}{d}}  \, .$$ 
So the bound on robustness improves  as the quantity
$$\frac{2r^2 \ln n}{d} - a(x)^2$$
becomes more negative.
Note that the dependence of this robustness bound on the number of network units is only  $\ln n$, 
whereas the dependence $d$ is stronger, scaling as $\frac{1}{d}$.

In practical applications, networks are typically of moderate size, making it reasonable to assume that the total number of units grows at most polynomially with $d$. Assuming that $n=d^k$, the condition characterizing the vulnerable neighborhood of the decision boundary becomes 
$$a(x)^2 < \frac{2r^2 k \ln d}{d}.$$
So the neighborhood of $H$, in which inputs may be susceptible to misclassification, is of order 
$\sqrt {\frac{\ln d}{d} }$ which srinks as the dimension $d$ grows.  For a fixed $d$ and a perturbation radius $r$, one can derive a lower bound on the distance $a(x)$ of an input $x \in \mathbb{R}^d$ from $H$ that guarantees that the probability of misclassification remains below a prescribed threshold. Equivalently, for any desired upper bound on the misclassification probability and any fixed perturbation radius $r$, one can determine a minimum distance from $H$ such that the parameterized ReLU network achieves the required level of robustness for all inputs $x$ with $a(x)$ above this threshold.

\begin{figure}[ht]
 \centerline{\includegraphics[width=12cm]{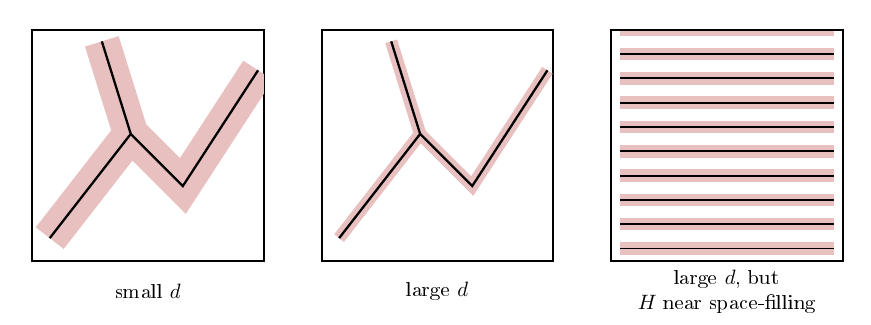}}
 \label{fig:spacefill}
\caption{Sets of points vulnerable to misclassification} \end{figure}

 Although the set $H$ has Lebesgue measure zero and can be described explicitly by computing the parameters $w_{l,j}$ and $c_{l,j}$ of the hyperplanes $H_{w_{l,j},c_{l,j}}$ constructed in the proof of Theorem~\ref{th:ReLU}, the volume of the neighbourhood of $H \cap X$ (where $X \subset \mathbb{R}^d$ is bounded), consisting of all points within the distance
$$ \sqrt{ \frac{2r^2 \ln n}{d}},$$
 \n depends on the geometry of $H$, which can be highly convoluted (see Fig. \ref{fig:spacefill}).
 
\n Note that ReLU networks realize shatterings on finite domains of size $m$ characterized by the growth function $\Pi(m)$ satisfying
$$ \Pi(m) \leq (4eLm)^{LW},$$ which grows exponentially with the network depth $L$ and the total number of parameters $W$ (see the proof of Theorem~7 in \cite{baal19} and the discussion on bound tightness). 
Nevertheless, in practical applications, the decision boundary $H$ is unlikely to exhibit a space-filling geometry. Consequently, as the input dimension $d$ increases, the proportion of vulnerable points can be expected to decrease.

\section{Conclusion}
\label{sec:disc}

We theoretically investigated the susceptibility of neural networks to average 
$\mathcal{L}_2$-perturbations by analyzing local robustness, quantified as the probability that a randomly perturbed input is not misclassified. We derived lower bounds on local robustness in terms of the input dimension, the perturbation size, and the distance to the boundaries of the convex cells in the partition of the input space induced by the I/O functions of ReLU networks. Our analysis combines concentration phenomena on high-dimensional spheres with a new estimate of the number of faces of each polyhedral region in this partition. In particular, we proved that although the number of regions may grow exponentially with network depth, each individual region is bounded by at most as many faces as there are network units. Our results imply that the local robustness at an input $x \in \mathbb{R}^d$ depends on the quantity $\frac{2r^2 \ln n}{d} - a(x)^2$, where $n$ denotes the total number of network units, $r$ the perturbation radius, and $a(x)$ the distance of input $x$ to the boundary set $H$.

Our proofs rely on the piecewise-linear structure induced by ReLU activations combined with tools from high-dimensional geometry. Consequently, our techniques do not directly extend to kernel methods or convolutional neural networks, whose robustness properties would require substantially different analytical tools.

We were motivated by the hypothesis of Szegedy et al.~\cite{szal14}, quoted in the Introduction. Concerning the second part of their hypothesis, namely that adversarial examples may form a dense set (analogous to the rational numbers), our analysis shows that this cannot hold for ReLU networks. As $H$ is closed and has Lebesgue measure zero, there are abundantly many points $x$ and distances $r$, for which the balls of radius $r$ centered at $x$ do not intersect $H$. For such $x$, there are no misclassified inputs arbitrarily close to $x$. Our results also provide partial support for the first part of their hypothesis, namely that adversarial examples occur with low probability. We proved that if the total number of ReLU units grows at most polynomially with the input dimension $d$, then the width of the neighborhood of the boundary set $H$ containing inputs susceptible to misclassification decreases of the order of $\frac{\ln d}{d}$. Consequently, unless $H$ exhibits an approximately space-filling geometry, the probability that a randomly sampled input lies outside this narrow neighborhood is expected to increase with the input dimension. 

In this paper, we focused on average-case adversarial robustness in the binary classification setting. Extending our analysis to worst-case robustness and multi-class classification with softmax outputs is a direction for our future work.

 \section*{Acknowledgments}
This work was partially supported by the Czech Science Foundation grant 25-15490S and the
institutional support of the Institute of Computer Science RVO 67985807.

\end{document}